# Transfer Learning for Melanoma Detection: Participation in ISIC 2017 Skin Lesion Classification Challenge

Dennis H. Murphree
Department of Health Sciences Research
Mayo Clinic
Rochester, MN
murphree.dennis@mayo.edu

Che Ngufor
Department of Health Sciences Research
Mayo Clinic
Rochester, MN
ngufor.che@mayo.edu

*Abstract*—This manuscript describes our participation in the International Skin Imaging Collaboration's 2017 Skin Lesion Analysis Towards Melanoma Detection competition. We participated in Part 3: Lesion Classification. The two stated goals of this binary image classification challenge were to distinguish between (a) melanoma and (b) nevus and seborrheic keratosis, followed by distinguishing between (a) seborrheic keratosis and (b) nevus and melanoma. We chose a deep neural network approach with a transfer learning strategy, using a pre-trained Inception V3 network as both a feature extractor to provide input for a multi-layer perceptron as well as fine-tuning an augmented Inception network. This approach yielded validation set AUC's of 0.84 on the second task and 0.76 on the first task, for an average AUC of 0.80. We joined the competition unfortunately late, and we look forward to improving on these results.

## I. Introduction and Dataset

In 2017 the International Skin Imaging Collaboration (ISIC) organized a three part competition geared towards skin lesion analysis, with a focus on melanoma detection. There were three parts to this competition – Lesion Segmentation, Lesion Dermoscopic Feature Extraction, and Lesion Classification. All of these parts revolved around analyzing dermatoscope images with an ultimate goal of enhancing the identification of melanoma. Our group participated in the third part of this competition, Lesion Classification. Here we describe our experiments using convolutional neural networks in two separate binary image classification tasks: distinguishing melanoma from nevus and seborrheic keratosis (Task 1), and distinguishing seborrheic keratosis from nevus and melanoma (Task 2). The dataset [1] provided by ISIC consisted of 2000 training set dermatoscope images, 150 validation set images, and 600 final test set images. Each image is accompanied by a "ground truth" diagnosis as determined by an unspecified number of skin cancer experts. In addition, each image is accompanied by a set of clinical metadata, specifically the age and sex of the patient. The images are RGB color and vary in size, ranging from 750,000 to as large as 30 million square pixels. For Task 1 the minority class (melanoma) was present in 19% of the cases, while in Task 2 the minority class (seborrheic keratosis) was present in 13%. Addressing this mild but nonetheless non-trivial class imbalance was discovered to be important during training.

We attempted four general network architectures, with three implemented by the end of the competition. These were a deep convolutional neural network (CNN) trained from scratch (Scratch), a transfer learning approach using an ImageNet pre-trained Inception v3 network [2] as a feature extractor to produce inputs to a two layer perceptron (FeatureExtractor), the same network fine-tuned with a lower learning rate (FineTune), and finally a hybrid network (Hybrid) which merged a CNN operating on the dermatoscope images with a three layer perceptron operating on the clinical metadata. The fourth network was only partially implemented by the competition end, and we mention it here solely for the purpose of receiving feedback from the community. The Scratch network substantially underperformed the transfer learning networks, but we include it here as an example of something which did not work well as well as for the purposes of receiving feedback from the community.

This clearly shortened work proceeds as follows: In Section II we describe the preprocessing steps performed prior to training our networks. Section III describes the networks, their training, and their results, and we conclude in Section IV.

## II. Preprocessing

### A. Image Resizing and Scale Conversion

For the Scratch network, all images were resized to 128x128 pixels then converted to grayscale via the 0.298839*R+0.586811*G+0.114350*B formula using ImageMagick's *convert*. The same conversion process was performed prior to training the FeatureExtractor and FineTune networks but resizing the raw images to 299x299. We also tried training the Scratch network without the conversion to grayscale and found no performance difference. We caution the reader however that the Scratch network's performance was poor, so a point for further experiment would be to measure any effect on the FeatureExtractor and FineTune networks. We also mention in passing that further rescaling the images by subtracting a global mean, a strategy suggested in [3], did not improve the performance of the Scratch network. As with the grayscale however due to time constraints we did not repeat the experiment on the better performing transfer learning networks.

*B. Minority Class Oversampling*

Oversampling is a common practice when developing predictive models on data with class imbalances largely because it is easy to implement and frequently improves models' predictive performance [4]. In particular, when large class imbalances are present in a dataset, naïve training of predictive models will often not be sensitive enough to the minority class. We found that models trained without oversampling did not produce competitive results. We recognize that many authors [5-8] have found that while most resampling methods are effective at dealing with class imbalance, the choice of resampling method is often both domain and problem specific, with no single best practice approach. While we could have tried a variety of data augmentation approaches, including well known shear, rotation and translation warping pathways, due to time limitation we took the time-parsimonious approach of simply duplicating each example of the minority class in our training set three times. The validation and test sets were not affected. Like the grayscale conversion experiment above, results with and without minority class oversampling were only compared using the Scratch model. With the Scratch model we found that a native class distribution was untrainable – the network simply always predicted the majority class. However, after introducing this triplicate multiplication of the minority class, the Scratch network was able to distinguish between the classes, albeit at an AUC of 60%. As with the grayscale experiment, it would be a useful next step to evaluate the effect of this oversampling on a higher performance network such as our FeatureExtractor or FineTuning nets. We emphasize that the oversampling was only applied to the training dataset, thus did not affect any hold-out evaluation. Similarly oversampling the training data did not affect any results on the competition's validation or final test sets.

### III. NETWORK ARCHITECTURE, TRAINING AND RESULTS

*A. Architecture and Training*

As mentioned in Section I we attempted four distinct neural network architectures: Scratch, FeatureExtractor, FineTuning, and Hybrid. We used the Keras framework [9] to build and train the models. All were trained with a *batchSize* of 32 and used a class-stratified validation set to evaluate performance over the course of training. Specifically, the data were divided into the following sets: training (67.5%), validation (7.5%), and final test (15%). Splits were performed in a class stratified way, and the oversampling routing mentioned above was only employed after splitting off the training set.

Scratch is a fairly generic deep learning convolutional neural network. Its architecture can be seen in Figure 1. We trained it on 128x128 grayscale images using stochastic gradient descent with parameters *learningRate*=0.001, *decay*=1e-6, *momentum*=0.9, and *nesterov*=True. The loss function was *binary_crossentropy* and the training optimization metric was accuracy. Scratch's results on both our own 10% validation set as well as the contests' provided validation set demonstrated AUC's in the mid 0.60's. Recalling that an AUC of 0.50 is random chance, we rapidly moved to a new approach.

For FeatureExtractor, we took the well known Inception V3 architecture [2] and replaced the final fully connected layer with a global average pooling layer followed by a fully connected layer with 1024 nodes and activated with ReLU. We then added a second fully connected layer with a single node output, activated by a sigmoid function. All layers other than the newly added ones were then frozen and the network was trained for a potentially short 20 epochs using *rmsprop* as the optimizer, *binary_crossentropy* as the loss, and *accuracy* as the optimization metric. We built and trained this model using the exceptionally helpful Keras framework [9], and incorporated an adaptive learning rate into the training which reduced the learning rate by a factor of 10 after five epochs of no improvement in the accuracy on the validation set.

For FineTuning, we used the best model found under the FeatureExtractor, but then modified it by un-freezing the last two Inception blocks. We then trained these two Inception blocks along with our final two fully connected layers. Instead of using *rmsprop* however we optimized via stochastic gradient descent with a learning rate of 1e-04, momentum of 0.9, *binary_crossentropy* for the loss function, and accuracy as the validation metric. For the FineTuning model we used both an early stopping criterion and an adaptive learning rate. They were configured as follows (Keras syntax):

earlyStopping = EarlyStopping(monitor='val_acc', min_delta=0.01, patience=50, verbose=1, mode='auto')

reduceLR = ReduceLROnPlateau(monitor='val_acc', factor=0.1, patience=25, verbose=1))

The Hybrid network architecture sought to combine the FineTuning network from above with a multi-layer perceptron accepting clinical metadata as inputs. The idea was to incorporate both image data as well as clinical data into a single predictive network. Although we feel that this is a worthy avenue of pursuit for future experiments, the Hybrid network's implementation wasn't completed by the competition end date. We look forward to improving it.

*B. Results*

Results as evaluated on the holdout test set can be seen in Table 1. While the high accuracies are not suprising given the class imbalance, we were a little surprised by the relatively high AUC's, and we welcome any comment from the community regarding this.

**Table 1: Test Set Results**

|  | Task1 AUC | Task1 Test Acc | Task1 Avg Prec | Task2 AUC | Task2 Test Acc | Task2 Avg Prec |
|---|---|---|---|---|---|---|
| Scratch | 0.62 | 0.77 | 0.24 | 0.73 | 0.81 | 0.49 |
| Feature Extractor | 0.77 | 0.80 | 0.39 | **0.87** | **0.92** | 0.60 |
| FineTune | **0.78** | **0.83** | **0.43** | 0.87 | 0.92 | **0.66** |

IV. CONCLUSION

In summary, our methods produced competitive but not superlative results. Next steps of interest include finishing the implementation of the hybrid network to include the clinical metadata, as well as exploring other "base" network architectures to try a more effective weight transfer. For example, preliminary results suggest that starting with AlexNet rather than Inception might prove a better choice. We are also interested in further tuning a variety of hyperparameters related to training such as batch size and initial image size, as well as parameters of our additional MLP network such as number of hidden layers, number of hidden nodes, and dropout.

We thank the organizers of the ISIC 2017 Challenge – we very much enjoyed participating, and we look forward to next year.

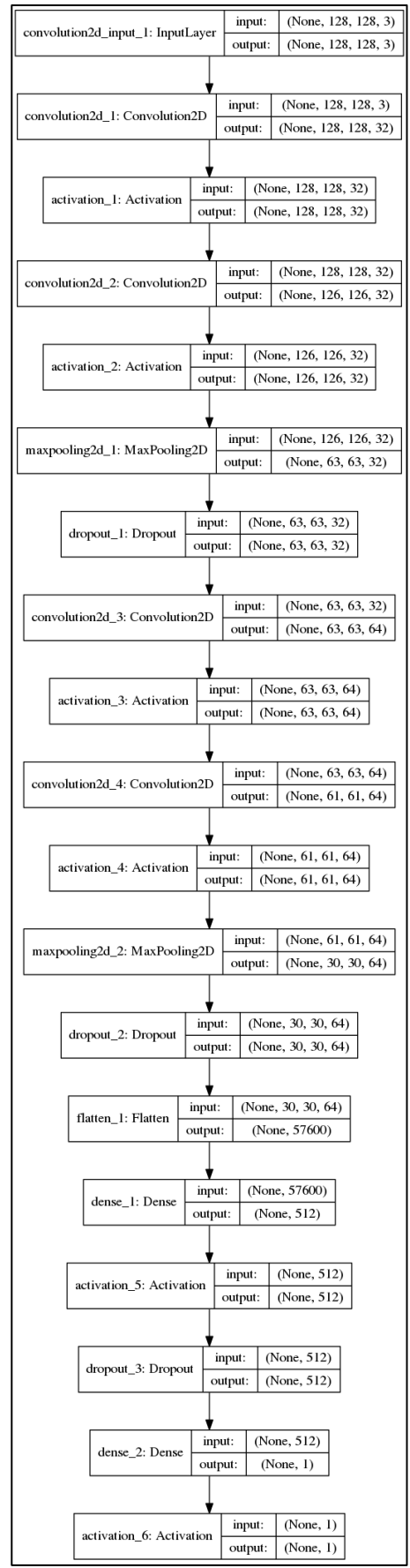

**Figure 1 – Scratch Model Architecture**